\documentclass{rsos}
\usepackage{booktabs}
\usepackage{graphicx}
\usepackage{amsmath}
\usepackage{algorithm}
\usepackage{algpseudocode}
\usepackage{xcolor}
\Year{2026}
\hypersetup{hidelinks, pdfauthor={Peter Chen, Ni Zhang, Siyuan Zhang}, pdfcreator={Peter Chen, Ni Zhang, Siyuan Zhang}}
\graphicspath{{figs/}}
\titlehead{Research}
\jname{rsos}
\Journal{R. Soc. Open Sci.\ }

\begin{document}

\title{Verifier strength does not buy discovery quality: a pre-registered
two-domain test of LLM-based automated research, with post-hoc evidence
pointing to proposer-pool composition}

\author{Peter Chen$^{1,\dagger}$, Ni Zhang$^{1,\dagger}$
and Siyuan Zhang$^{1,*}$}

\address{$^{1}$Shenyang Institute of Technology, Shenyang, China}

\subject{computer science; artificial intelligence; statistics}

\keywords{automated research; LLM proposers; pre-registration; verifier
strength; proposer-pool composition}

\corres{Siyuan Zhang\\
\email{lelandsy28@163.com}}

\others{$^{\dagger}$These authors contributed equally to this work.}

\begin{abstract}
Automated research pipelines pair an LLM proposer with a verifier on an
implicit bet: discovery quality rises with verification strength. Under a
pre-registered protocol we test this bet in two real-data domains --
job-shop scheduling and vehicular edge offloading on Beijing T-Drive
trajectories -- across 100 matched LLM/random pairs per domain and three
verifier tiers. The manipulation binds -- a 429-case mutation battery shows
scores rising across tiers -- yet the paired LLM-over-random
advantage does not scale with it (A, one-sided $p{=}0.1115$; D,
$p{=}0.9994$), and in D shrinks under the strict tier (margin contrast
$-0.040$, $p{=}0.0013$). The two lower gates promote identical candidates
in 100/100 and 99/100 pairs, so the strict order is untestable when they
coincide. Constructively, LLM proposals carry a real paired advantage over
random controls, but a post-hoc union-pool analysis shows the fastest path
to better discovery is widening the pool, not tightening the gate: mixing
the two pools adds $+1.52$ holdout-value units in domain A (one-sided
Wilcoxon, $n{=}100$, $p{=}0.0002$) -- evidence that pool composition, not
verification strength, is the lever. Criteria, ledgers and scripts are
archived on Zenodo (DOI: 10.5281/zenodo.23308866).
\end{abstract}

\maketitle

\section{Introduction}
Automated scientific discovery pipelines built on large language models
\cite{lu2024aiscientist} increasingly pair an LLM proposer with an automated
verifier: the model generates candidate decision rules, heuristics, or
hypotheses; a programmatic verifier screens them against constraints; the
surviving candidate is scored on held-out data. The scaling behavior of such
pipelines is usually discussed in terms of proposer capability. There is,
however, a second lever that is cheaper to turn and is widely assumed to
monotonically improve outcomes: \emph{verification strength}. If screening is
what stands between noisy proposals and discovery quality, then a stricter
verifier -- more constraints, plus adversarial stress testing -- should buy
more quality. This assumption shapes how practitioners build such systems:
add checks, add stress tests, trust the filter.

This paper tests the assumption directly rather than assuming it. We decompose
a minimal automated-research pipeline into a proposer $P$ (an LLM under
iterative verifier feedback) and a verifier $V$, and ask a causal question:
\emph{holding $P$ fixed, does increasing the strength of $V$ improve the
quality of what is discovered?} Verification strength is operationalized as
three frozen tiers: S$^-$ (minimal screening), S$^0$ (the full constraint
verifier), and S$^+$ (S$^0$ plus a worst-case stress test). Discovery quality
is the one-shot held-out value of the promoted rule under a frozen scoring
function.

Answering this credibly requires three design commitments. First,
\emph{pre-registration}: the criteria, the protocol, and the analysis were
frozen (v1.0) before any main-domain data were collected, with changelogs
disclosing every deviation, including a manipulation-check population
disclosure (v1.2) made before unblinding. Second, a \emph{manipulation check
before the main data}: a mutation battery must establish that the three tiers
genuinely differ in what they filter (E1-D2); a main experiment with an
ineffective manipulation is uninterpretable. Third, a \emph{matched control
arm}: every LLM session is paired with a random-proposal session of equal
size, so that quality is measured as the paired advantage of structured
proposals over random ones, not as an absolute score.

We ran 100 such pairs in each of two real-data domains: job-shop scheduling
on instances derived from real records (domain A), and vehicular edge
offloading on Beijing T-Drive taxi trajectories \cite{yuan2010tdrive}
(domain D). The proposer is a small open-weights model
(Qwen2.5-0.5B-Instruct \cite{qwen25}) running fully offline on CPU -- a
deliberately modest proposer, so that any conclusion applies to the
pipeline structure rather than to frontier-model capability, and so that the
whole study is reproducible on commodity hardware.

The main hypothesis fails in both domains. The manipulation check passes
(the 429-case mutation battery yields scores that strictly increase
across tiers), yet the paired quality
advantage of LLM proposals over random ones does not increase with verifier
strength: it is statistically indistinguishable across tiers in domain A and
\emph{decreases} under the strict tier in domain D. The failure is
informative rather than null. Three additional analyses give it structure:
(i) the intermediate feasibility gate is \emph{inert} -- in 100/100 and
99/100 pairs S$^-$ and S$^0$ promote the identical candidate, because an
LLM session's proposals are near-determinately all-feasible or
all-infeasible; (ii) the LLM pool has a large \emph{coverage gap} -- the
best of twenty random proposals beats the best of twenty LLM proposals in
roughly half the pairs, and adding the random pool would have significantly
improved held-out quality in domain A ($+1.52$, $p{=}0.0002$), a gain that
stricter verification failed to deliver; (iii) a post-hoc \emph{tension
gradient} shows the value of the stress-test tier crossing zero with domain
tension -- negative when the deployment regime leaves constraints slack,
peaking at medium utilization. Together these suggest a separation result:
verifier tiers that separate cleanly on the mutation battery do not buy
discovery quality; the post-hoc evidence points instead to proposer-pool
composition as the lever, and that evidence is now under pre-registered
test as an intervention.

We therefore close with two pre-registered follow-ups that treat this
conclusion as an experimental hypothesis: a diversity dose--response
experiment (pool composition manipulated at fixed verifier) and a tension
transfer experiment (utilization manipulated at fixed verifier). All
criteria, ledgers, and seed-to-figure reproduction scripts are archived
on Zenodo (DOI: 10.5281/zenodo.23308866).

\paragraph{Contributions.} This paper makes four concrete contributions:
\begin{enumerate}
\item \textbf{A pre-registered separation result} (Sections~\ref{sec:setup}%
--\ref{sec:gate}): with the manipulation check passing and the main
hypothesis failing under the same frozen criteria, verifier strength is
decoupled from discovery quality in two real-data domains -- the
cleanest null this design can produce.
\item \textbf{A quantified coverage gap} (Section~\ref{sec:gate}): to our
knowledge the first measurement of the LLM-vs-random best-of-$n$ gap
inside a verifier-bound discovery pipeline, showing the best of twenty
random proposals beats the best of twenty LLM proposals about half the
time, and a derived (post-hoc) union-pool estimate of what widening the
pool would buy ($+1.52$ in domain A).
\item \textbf{A criterion-reachability lesson for pre-registration}
(Section~\ref{sec:gate}): a strict-order criterion is structurally
untestable when two treatment arms coincide -- reachability of criteria
under plausible degenerate outcomes must be checked before freezing.
\item \textbf{Two pre-registered follow-ups} (Section~\ref{sec:followups})
that convert the post-hoc findings into testable
interventions: diversity dose--response at fixed verifier, and tension
transfer at fixed tier.
\end{enumerate}

\paragraph{Paper roadmap.} Section~2 locates the contribution relative to
automated-research systems, LLM-as-judge and verifier literature, and
quality--diversity methods. Section~3 states the framework and frozen
criteria. Section~4 describes domains, protocol, and the manipulation
battery. Section~5 reports the manipulation check, the main result, the
gate-degeneracy and coverage-gap analyses, and the tension gradient.
Section~6 discusses implications; Section~7 lists limitations; the back
matter covers reproducibility, data availability, ethics, and declarations.

\section{Related work}
\paragraph{Automated research systems} End-to-end systems such as The AI
Scientist \cite{lu2024aiscientist} chain an LLM proposer with automated
execution and evaluation, and evaluate the output with human or AI review;
domain pipelines have produced autonomous chemistry laboratories
\cite{boiko2023coscientist} and mathematical conjectures via program search
\cite{romera2024funsearch}, and a large-scale human study finds LLM research
ideas rated as novel while LLM self-evaluation remains unreliable
\cite{si2025ideas}, and a recent survey argues the field's bottleneck is
shifting from task completion to verifying what agents claim
\cite{ding2026survey}.
Our contribution is orthogonal: we hold the proposer fixed and manipulate
the verifier, under pre-registration, to isolate the causal contribution of
verification strength -- making that proposed bottleneck directly measurable. The proposer-heavy line treats filtering as plumbing;
we show the plumbing is where a central assumption fails.

\paragraph{LLM-as-judge and programmatic verifiers} The alignment
literature studies whether LLM judges match human preferences
\cite{zheng2023judging}. Our verifiers are programmatic and exact
(constraint checkers, stress tests), removing judge-noise as an explanation:
when quality fails to track verification strength here, it is not because
the verifier is noisy. Selection pipelines that assume stronger filtering
buys quality -- outcome and process verifiers
\cite{cobbe2021verifiers,lightman2024lets}, self-consistency
\cite{wang2023selfconsistency}, and verbal self-refinement
\cite{shinn2023reflexion,madaan2023selfrefine} -- inherit precisely this
assumption, and resampling against imperfect verifiers is already known to
be capped by false positives \cite{stroebl2026limits}; our randomized
evidence extends that concern from pass-rate contamination to discovery
quality. The generator--verifier gap literature
\cite{brown2024monkeys} shows pass-rate gaps between generation and
verification; we show a complementary gap -- between verification strength
and downstream quality. Two concurrent audits in adjacent domains reinforce
this audit premise: agents answering scientific questions often ignore the
supplied evidence \cite{zhang2026measuring}, and LLM-generated kernels that
pass numerical checks can fail under scale shift \cite{suwandi2026kernaut}.
Neither manipulates verifier strength under pre-registration or measures
discovery quality, which is our focus.

\paragraph{Pool composition and diversity} Quality-diversity methods
\cite{mouret2015elites,pugh2016qd}, descendants of novelty search
\cite{lehman2011novelty}, make archive diversity the explicit optimization
target; repeated-sampling results \cite{brown2024monkeys} show coverage
growing steeply with draw count. Adjacent to our negative result, a
budget-sensitive scoring study finds no marginal discovery value from LLM
proposers over selection alone in a drug-candidate pipeline
\cite{basu2026budget}. Our union-pool analysis is, to our
knowledge, the first to quantify the LLM-vs-random coverage gap inside a
verifier-bound discovery pipeline, and the first to show it widening under
stricter verification.

\paragraph{Pre-registration and negative results} Our protocol follows
clinical-trial norms transplanted to ML evaluation: frozen criteria with
changelogs, a manipulation check gate before the main data, and a negative
clause committing to publication of failure. These norms respond to
replication failures across disciplines \cite{osc2015science,%
errington2021replicability} and to reproducibility gaps specific to AI and
ML-based science \cite{gundersen2018state,pineau2021reproducibility,%
kapoor2023leakage}. The S$^0\equiv$S$^-$
degeneracy adds a criterion-reachability lesson to this practice: check
that a criterion can be satisfied under plausible degenerate outcomes
before freezing it.

\section{Framework and frozen criteria}
\subsection{Proposer--verifier decomposition}
\begin{sloppypar}
An automated-research pipeline is a proposer $P$ (an LLM under iterative
feedback, 20 proposals per session) and a verifier $V$ mapping a
candidate to \{pass, fail\} plus a frozen score. Discovery quality on a
held-out instance set $H$ is the holdout value of the promoted candidate
$\arg\max_{c:\,V(c)=\mathrm{pass}} \mathrm{train}(c)$, evaluated exactly
once on $H$ (HoldoutGate: one-shot semantics, no re-evaluation). We write
$J(c)$ for the frozen holdout score of candidate $c$ and
$\overline{J}(G)$ for its mean over a group $G$ of promoted rules
(used in Fig.~\ref{fig:tension}); $\Delta(\cdot)$ denotes the paired
LLM-over-random margin of a tier, and $\mathrm{MS}(\cdot)$ the mutation
score of a tier (Section~\ref{sec:setup}).
\end{sloppypar}

\begin{table}[t]
\centering\small
\caption{Notation summary.}
\label{tab:notation}
\begin{tabular}{ll}
\toprule
symbol & meaning \\
\midrule
$P$ / $V$ & proposer (LLM under iterative feedback) / verifier \\
S$^-$, S$^0$, S$^+$ & weak / standard / strict verifier tier (Sec.~\ref{sec:tiers}) \\
$J(c)$ & frozen holdout score of candidate $c$ \\
$\overline{J}(G)$ & mean of $J$ over a group $G$ of promoted rules \\
$\Delta(\cdot)$ & paired LLM-over-random margin of a tier \\
$\mathrm{MS}(\cdot)$ & mutation score of a tier (Sec.~\ref{sec:setup}) \\
$H$ & held-out instance set (one-shot evaluation) \\
RSU / tpv & road-side unit / tasks-per-vehicle (domain D) \\
\bottomrule
\end{tabular}
\end{table}

\subsection{Three verifier tiers}
\label{sec:tiers}
\begin{itemize}
\item \textbf{S$^-$} (weak): minimal screening. Domain A promotes any
  rule that executes (INFEASIBLE outcomes score the frozen floor
  $-10^9$); domain D promotes any rule without deadline violations.
\item \textbf{S$^0$} (standard): the full frozen constraint verifier
  (A: feasibility-gated scoring; D: safety isolation, per-task deadline,
  per-task energy allowance, per-RSU capacity).
\item \textbf{S$^+$} (strict): S$^0$ plus a worst-case stress test
  (A: non-degeneracy of promoted weights; D: re-verification with every
  task's channel quality lowered to the instance minimum).
\end{itemize}

\subsection{Frozen criteria (v1.0)}
\textbf{E1-D1 (main):} the paired advantage of the LLM arm over the
random control, $\Delta(\cdot)$, satisfies
$\Delta(\mathrm{S}^+) > \Delta(\mathrm{S}^0) > \Delta(\mathrm{S}^-)$
with one-sided Wilcoxon $p<0.05$ for S$^+$ vs.\ S$^-$, in both main
domains (domain A mandatory). \textbf{E1-D2 (manipulation validity):}
strictly ordered mutation scores $\mathrm{MS}(\mathrm{S}^-)<
\mathrm{MS}(\mathrm{S}^0)<\mathrm{MS}(\mathrm{S}^+)$ in both domains.
\textbf{Negative clause:} if either criterion fails, the manipulation is
recorded as ineffective or the main hypothesis as falsified, and all
outcomes are reported as such.

\section{Experimental setup}\label{sec:setup}
\subsection{Domains and protocol}
\textbf{Domain A (scheduling).} Priority rules over 7 normalized
numerical features of the job-shop instance (frozen feature list in the
archived protocol). Instances are bootstrap resamples of a real
production-order pool from the prior suture experiments of this program:
all listed orders in a fixed due window (2026-08-15 to 2026-09-02),
deduplicated across 10 business fields -- 1{,}791 rows collapse to 154
unique orders, because the same batch was re-imported under different run
identifiers. Pair $i$ (seed $1000{+}i$) draws a 60-order train instance
without replacement and a disjoint 60-order holdout instance from the
remaining 94 orders, redrawing if the baseline schedule is infeasible or
exceeds 40 days. The pool is small and pairs are strongly correlated by
construction, so domain-A claims are limited to within-distribution
perturbation robustness -- not across-factory generalization.
\textbf{Domain D (offloading).}
Offloading-ratio rules over 8 normalized task features; instances built
from Beijing T-Drive taxi trajectories \cite{yuan2010tdrive} around 3
Road Side Units (RSUs) by the frozen instance generator of the archived
pipeline: pair $i$ uses a frozen training-instance seed ($9000{+}i$) and
a disjoint holdout-instance seed ($10000{+}i$). T-Drive is a public
research dataset released by
Microsoft Research; trajectories were aggregated into per-pair task
arrival instances at frozen timestamps, with no identifier or
trip-purpose information retained (Section~\emph{Ethics}). Holdout
instances for both domains are held out by construction: frozen at
protocol time, never scored during training, and evaluated exactly once
per promoted rule.

\paragraph{Session and gating design.} Each domain runs 100 matched
pairs. Per pair: \emph{one} fresh LLM session (20 iterative proposals
under standard-tier S$^0$ prompt and verifier feedback) and one random
control (20 uniform weight vectors over the same rule family, per-pair
seed). The \emph{same} 20-proposal pool is then re-gated independently by
each verifier tier -- S$^-$, S$^0$, S$^+$ -- and each tier promotes the
passing candidate with the best training value; the random pool is gated
identically. Tiers therefore differ \emph{only} in the filter applied to
a fixed proposal pool, which is what makes the manipulation a clean
test of verification strength. Promotion gaps (an arm has no passing
candidate in a tier) occurred in 0/100 pairs in domain A and in 3 pairs
in domain D; those 3 pairs are excluded from analyses involving the
affected arm (n=99/97 where applicable, stated per analysis).

\begin{algorithm}[t]
\caption{Tier-1 paired session with three-tier re-gating (per pair)}
\label{alg:pipeline}
\begin{algorithmic}[1]
\State draw frozen train instance and holdout instance (frozen per-pair seeds)
\State \emph{LLM arm}: run one fresh session of 20 iterative proposals
  under the S$^0$ prompt and verifier feedback
\State \emph{random arm}: draw 20 proposals from the frozen random
  distribution (per-pair seed, same budget)
\For{each tier $T \in \{\mathrm{S}^-, \mathrm{S}^0, \mathrm{S}^+\}$}
  \State gate \emph{both} fixed proposal pools by tier $T$
  \State promote the passing candidate with the best training value
\EndFor
\State evaluate each promoted rule on the holdout instance exactly once
  (HoldoutGate: one-shot semantics)
\State \textbf{return} per-tier promotions and holdout values
\end{algorithmic}
\end{algorithm}

\paragraph{Proposer and compute.} Proposer: Qwen2.5-0.5B-Instruct
\cite{qwen25} (0.49B parameters, 32K context), served locally through
Hugging Face \texttt{transformers} on PyTorch with CPU inference, fully
offline (hub and local-restore environment flags set). Decoding is
frozen: sampling with temperature $0.8$, top-$p$ $0.9$; up to 220 new
tokens per proposal in domain A (Chinese prompt template with two fixed
few-shot examples) and 180 in domain D (template with
per-category verifier feedback); sampling is not seeded, so the unit of
replication is the 20-proposal session, and the archived ledgers record
every proposal verbatim. The frozen prompt and generation configuration
are reproduced verbatim in the archived protocol and ledgers. All
experiments ran offline on a consumer laptop CPU (no GPU); per-session
LLM wall-clock is minutes, and domain D is dominated
by trajectory simulation (on the order of half a day for the full
100-pair run).

\subsection{Manipulation battery (E1-D2)}
Domain A: 119 mutants (100 uniform random over the frozen 7-dim family,
19 structured). Domain D: 310 mutants (300 random rule templates, 4
extreme heuristics, 6 structured mutants matched to the violation mix
observed in a prior instrumented ledger: safety/energy/capacity entries
vs.\ deadline). Population disclosed before unblinding (v1.2).

\section{Results}
\subsection{Manipulation validity holds (E1-D2)}
\begin{table}[t]
\centering\small
\caption{Manipulation check (E1-D2): mutation score per tier
(fraction of battery mutants killed), with Wilson 95\% CIs. Populations
disclosed before unblinding (protocol v1.2).}
\label{tab:d2}
\begin{tabular}{lrrrr}
\toprule
domain & $n$ & MS(S$^-$) & MS(S$^0$) & MS(S$^+$) \\
\midrule
A & 119 & 0.000 [0.000, 0.031] & 0.0168 [0.005, 0.059] & 0.1092 [0.065, 0.178] \\
D & 310 & 0.3484 [0.298, 0.403] & 0.3548 [0.304, 0.410] & 0.3613 [0.310, 0.416] \\
\bottomrule
\end{tabular}
\end{table}
Mutation scores increase strictly in both domains
(Table~\ref{tab:d2}). The three tiers are genuine, ordered treatments.
In domain A the gaps are large relative to the CIs; in domain D the
ordering is strict but the gaps are small -- adjacent tiers differ by
2 of 310 mutants (0.3484 vs.\ 0.3548 vs.\ 0.3613, overlapping Wilson
CIs) -- so we describe domain D as \emph{strictly ordered but weakly
separated}, and rely on the domain-A battery, where the tiers separate
cleanly, as the primary manipulation evidence.

\subsection{Main result: discovery quality does not follow verification
strength (E1-D1 fails)}
\begin{table}[t]
\centering\small
\caption{Main result. $\Delta$(tier) = holdout value of the LLM-promoted
rule minus that of the random control (frozen scoring). $p$: one-sided
Wilcoxon, S$^+$ vs.\ S$^-$ (the frozen criterion). $\Delta$ scales are
domain-specific (different objectives) and are \emph{not} comparable
across rows. SD: per-tier margin standard deviation -- floor-free in
domain D ($0.19$--$0.22$, matching the bootstrap CIs in
Table~\ref{tab:ds}); in domain A the raw SDs are meaningless because of
the $-10^9$ floor contamination, so floor-free means on $n{=}98$ pairs
(Section~5.2, sensitivity analysis) are reported instead. S$^0\equiv$S$^-$
degeneracy makes the S$^0$-vs-S$^-$ and S$^+$-vs-S$^0$ contrasts
untestable (identical promotions).}
\label{tab:main}
\begin{tabular}{lrrrrr}
\toprule
domain & $\Delta$(S$^-$) & $\Delta$(S$^0$) & $\Delta$(S$^+$) &
SD & $p$ \\
\midrule
A & $+0.28$ & $+0.28$ & $+0.38$ & -- (floor) & $0.1115$ \\
D & $+0.10$ & $+0.10$ & $+0.06$ & $0.211/0.211/0.193$ & $0.9994$ \\
\bottomrule
\end{tabular}
\end{table}
The strict order fails in both domains (Table~\ref{tab:main}). Per-tier
post-hoc paired tests size the LLM-vs-random advantage directly
(one-sided Wilcoxon, LLM $>$ random; post-hoc, not part of the frozen
criteria): in domain D the advantage is significant at every tier
(S$^-$: $p{<}0.0001$; S$^0$: $p{<}0.0001$; S$^+$: $p{=}0.011$) yet
\emph{shrinks} monotonically with verifier strength; in domain A it is
positive at every tier ($+0.28/+0.28/+0.38$) but not significant at
$n{=}100$ ($p{=}0.31/0.31/0.24$; floor-free sensitivity below). The
frozen S$^+$-vs-S$^-$ contrast is far from significance in domain A, and
in domain D it points the \emph{wrong} way: $p{=}0.9994$ implies
$p{\approx}0.0006$ for the reverse ordering, and the post-hoc margin
contrast below is significantly negative (two-sided $p{=}0.0013$).
\textbf{The main hypothesis is falsified under the frozen criteria, and
the falsification is not a plain null: in domain D, stricter active
verification measurably reduced the LLM-over-random advantage.}

Two floor-contaminated pairs in domain A (an INFEASIBLE rule's value is
floored at $-10^9$ in both training and holdout; in domain A the floor is
a \emph{scored outcome} -- an executing rule whose schedule is infeasible
-- so both gated tiers can promote it) are excluded in a sensitivity
analysis re-run by \texttt{floor\_sensitivity.py}: on the $n{=}98$
floor-free pairs, $\Delta(\mathrm{S}^-){=}{+}0.367$,
$\Delta(\mathrm{S}^+){=}{+}0.469$, and the one-sided Wilcoxon $p$ is
unchanged ($0.1115$). The small raw mean and the unchanged $p$ both
follow from how the two floor pairs enter: in one the \emph{random}
arm's promoted rule is INFEASIBLE (a $+10^9$ margin) and in the other the
\emph{LLM} arm's is (a $-10^9$ margin), so the two cancel in the raw mean
-- leaving $\Delta{=}{+}0.28$ while the SD is floor-dominated at
$1.4\times10^8$ -- and they contribute zero to the signed-rank statistic
because $\mathrm{S}^+{=}\mathrm{S}^-$ within each floor pair. Excluding
them moves the means ($+0.367/+0.469$) but not the test ($0.1115$). No
domain-D pair involves the floor.

\begin{table}[t]
\centering\small
\caption{Per-tier LLM-over-random margins with bootstrap 95\% CIs
(post-hoc, computed by \texttt{descriptive\_stats.py}; not part of the
frozen criteria). Domain A is reported on the $n{=}98$ floor-free
pairs; domain D on its $n{=}97$ valid pairs.}
\label{tab:ds}
\begin{tabular}{lrrr}
\toprule
domain & $\Delta$(S$^-$) & $\Delta$(S$^0$) & $\Delta$(S$^+$) \\
\midrule
A (n=98) & $+0.367$ $[-1.15,\,1.85]$ & $+0.367$ $[-1.18,\,1.88]$
         & $+0.469$ $[-1.08,\,2.02]$ \\
D (n=97) & $+0.102$ $[+0.060,\,0.145]$ & $+0.102$ $[+0.061,\,0.145]$
         & $+0.062$ $[+0.024,\,0.100]$ \\
\bottomrule
\end{tabular}
\end{table}

Post-hoc descriptive statistics (not pre-registered criteria; computed by
\texttt{descriptive\_stats.py} under the frozen inclusion rule) size this
null. Two different contrasts appear here and must not be conflated:
the \emph{direct} S$^+$-vs-S$^-$ holdout difference of promoted rules,
and the \emph{margin contrast} -- the paired difference of the
LLM-over-random margins, $\left[\Delta(\mathrm{S}^+)-\Delta(\mathrm{S}^-)\right]$,
which subtracts off the random arm. The margin contrast has mean
$+0.10$ in domain A (95\% bootstrap CI $[-0.10,\,0.31]$,
two-sided $p{=}0.2229$) and $-0.040$ in domain D (95\% bootstrap CI
$[-0.07,\,-0.01]$, two-sided $p{=}0.0013$): domain D excludes zero
\emph{in the direction opposite to
the hypothesis}, i.e.\ stricter verification significantly reduced the
LLM-over-random margin there. (The smaller direct-difference means --
$+0.10$ in A over the 10 winning pairs and $-0.016$ in D over the
1/97 valid pairs -- quantify how rarely the strict tier changes the
promoted rule at all.) A power reading sharpens the asymmetry: at
$n{=}100$ (one-sided $\alpha{=}0.05$, power $0.8$), the minimum detectable
paired difference is $\approx 0.26$ in A, so a true positive effect up to
that size cannot be excluded; in D the observed $-0.040$ exceeds the
domain's MDE ($0.039$), so the negative association is detected rather
than missed. Per-tier margin SDs are $0.19$--$0.22$ in domain D; in
domain A the raw SDs are meaningless because of the $-10^9$ floor
contamination (floor-free values in the sensitivity analysis above).

\subsection{The feasibility gate is inert}\label{sec:gate}
\begin{table}[t]
\centering\small
\caption{Tier degeneracy and stress-test behavior. ``identical'': pairs
where S$^-$ and S$^0$ promote the same candidate. ``diff pairs'': pairs
where S$^+$ and S$^0$ promote different candidates, with S$^+$ wins and
mean holdout difference among those pairs.}
\label{tab:tiers}
\begin{tabular}{lrrrr}
\toprule
domain & identical & diff pairs & S$^+$ wins & mean diff \\
\midrule
A (n=100) & 100 & 14 & 10 & $+0.714$ \\
D (n=100) & 99 & 23 & 1 & $-0.066$ \\
\bottomrule
\end{tabular}
\end{table}
In \textbf{100/100} pairs (A) and \textbf{99/100} (D), S$^-$ and S$^0$
promoted the \emph{identical} candidate: proposals within an LLM session
are near-determinately all-feasible or all-infeasible, so an intermediate
gate almost never catches an intermediate candidate
(Table~\ref{tab:tiers}). The mechanism is exact in domain A: an
executing-but-infeasible candidate scores the frozen training floor
$-10^9$, which is never the session maximum unless \emph{every}
candidate in the session fails; under verifier feedback no session was
all-infeasible (0 promotion gaps in 100 pairs, both arms and all
tiers), so the best training-value candidate is always S$^0$-feasible
and both gates promote it. A session with no S$^0$-passing candidate
would leave S$^0$ without a promotion and break the degeneracy by
construction; the three domain-D promotion gaps are of this kind.
Two consequences: (i) the frozen strict order
S$^+>$S$^0>$S$^-$ is structurally unsatisfiable whenever
S$^0\equiv$S$^-$ (a criterion-reachability lesson for pre-registration
practice); (ii) in this regime the effect of verification strength is
carried entirely by the stress-test tier S$^+$ -- and that effect is positive in domain A
(S$^+$ wins 10 of its 14 divergent pairs) but sharply negative in domain
D (1 of 23). A concrete domain-D failure mode: in one pair the LLM
session passed 5 of 20 proposals under S$^0$, whose best candidate
scored $0.524$ on holdout; the S$^+$ stress test (worst-case channel
floor) killed two of the five passing candidates and forced promotion of
the second-best, which violated a constraint on the holdout instance and
scored $0.0$ -- a $-0.52$ swing produced purely by the stress test
rejecting the candidate that would have generalized. This is the
false-negative mechanism behind the domain-D margin shrinkage: in a
tense domain, the stress test removes correct-but-fragile-on-paper
candidates and the remaining pool is too thin to absorb the loss.

\subsection{The proposer pool has a large coverage gap}
Because each arm's tier promotion is the best passing candidate by train
value, the promotion of the union pool (LLM 20 + random 20 proposals) is
the better of the two arm promotions, and its holdout value is already
known -- the analysis is a zero-new-inference derivation (exploratory,
post-hoc). Three findings:
\begin{enumerate}
\item The union-pool promotion comes from the \emph{random} arm in 48\%
  of pairs (A) and 46\% (D) at S$^0$: the best of twenty LLM proposals
  is beaten by the best of twenty random draws about half the time.
\item Under S$^+$ (domain D) the random arm wins \textbf{60\%} of pairs:
  the LLM coverage gap \emph{widens} exactly where verification is
  strictest.
\item In domain A, adding the random pool to the LLM pool would have
  \emph{improved} holdout quality: $+1.52$ in domain-A
  holdout-value units (one-sided Wilcoxon signed-rank on per-pair
  holdout differences, $n{=}100$, $p{=}0.0002$) -- a gain that stricter
  verification failed to deliver. The same derivation at the strict
  tier gives $+1.44$ ($p{=}0.0006$), and in domain D the merge effect is
  null-to-negative ($-0.036$/$-0.041$/$-0.035$ across tiers,
  $p{=}0.27$/$0.32$/$0.07$): the pool lever has domain structure, which
  is why it is now the object of a pre-registered intervention (E2)
  rather than a settled claim. These are \emph{derived} statistics: the
  union-pool promotion is computed from the two arm promotions without
  new inference, and the union selection itself is realizable (best
  training value over the merged pool) -- what is post-hoc is our
  attention to it, not its implementability.
\end{enumerate}

\subsection{Post-hoc tension gradient: the S$^+$ effect crosses zero}
\begin{figure}[t]
\centering
\includegraphics[width=0.8\linewidth]{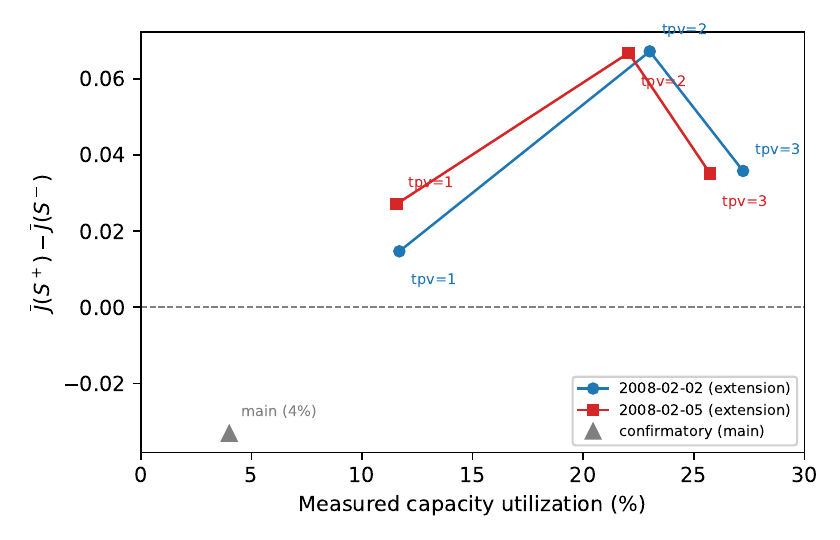}
\caption{Group contrast $\overline{J}(\mathrm{S}^+)-\overline{J}(\mathrm{S}^-)$
against measured capacity utilization ($J$: the frozen holdout score of a
promoted rule, Section~3.1; the figure legend abbreviates
tasks-per-vehicle as ``tpv''). The contrast crosses zero: negative in
the idle confirmatory regime (4\%), positive on both extension days, peaking
at medium utilization (22--23\%). Script-generated from
\texttt{tension\_gradient.json} and the frozen tier-1 ledgers.}
\label{fig:tension}
\end{figure}
Promoted-rule groups (first 50 pairs, LLM arm) were re-evaluated on 36
fresh instances spanning a capacity-utilization gradient (extension days
2008-02-02/05 $\times$ tasks-per-vehicle 1/2/3; measured utilization
12/22--23/26--27\%; the idle confirmatory regime of the main experiment sits at
4\%). The 50-pair subset is the first half of the frozen day-1 ledger in
collection order -- fixed before this analysis was unblinded to bound the
CPU cost of re-scoring every promoted rule on 36 fresh instances -- and
thus involves no outcome-based selection.
The group contrast $\overline{J}(\mathrm{S}^+)-
\overline{J}(\mathrm{S}^-)$ crosses zero with tension: $-0.033$ at 4\%,
$+0.015/+0.027$ at 12\%, $+0.067/+0.067$ at 22--23\%, $+0.036/+0.035$
at 26--27\% -- the same pattern on both extension days
(Fig.~\ref{fig:tension}). Stricter verification is not uniformly
helpful or harmful: its value is modulated by domain tension, turning
negative when the deployment regime leaves constraints slack.

\section{Discussion}
\paragraph{A separation result} The combination of E1-D2 passing and E1-D1
failing is the paper's central finding, and it is a separation result:
verification strength filters synthetic bad candidates -- the mutation
battery establishes that the tiers are real, ordered treatments -- while buying no
measurable discovery quality on LLM proposals. The three analyses of Section~5 explain why. First, the
feasibility gate is inert because proposals within an LLM session are
strongly correlated -- the model either understands the interface or it does
not, and within a session it almost always does one or the other
consistently. Intermediate constraints therefore almost never catch
intermediate candidates, and S$^0$ degenerates to S$^-$. Second, the
coverage gap means the marginal value of verification is bounded by what
the proposer pool contains: a filter, however strict, cannot promote a
candidate that was never proposed, and in these domains the best random
draw beats the best LLM draw about half the time. Third, the tension
gradient shows that even the one tier with a real effect (S$^+$) pays only
when the deployment regime makes its stress test bite; in a slack regime
the stress test is a source of false negatives, not quality.

\paragraph{Implications for AI-scientist pipelines} The results redirect
effort. The common escalation -- more constraints, stricter tests, more
adversarial screening -- targets the wrong bottleneck in this regime: the
feasibility gate is already inert, and the stress test helps only in tense
regimes. The first lever our evidence points to is \emph{pool
composition}: diversity of
generation (temperature, mixed LLM and random pools, multiple proposer
models), because the union-pool analysis shows random proposals would have
improved the merged promotion in domain A. We state the status of that
number precisely: the union selection rule (best training value over the
merged pool) is realizable without any holdout knowledge, and the
$+1.52$ is the holdout estimate of exactly that realizable policy,
computed post-hoc from already-evaluated arm promotions under one-shot
semantics. What it lacks is pre-registration as an intervention -- which
is precisely what the pre-registered E2 follow-up now supplies.
Verification
budget should follow deployment tension: strict tiers where constraints
bind, cheap tiers where they do not.

\paragraph{On the negative result} We report the main hypothesis as
falsified under frozen criteria, with the manipulation check passing --
the cleanest null this design can produce: the treatment is real, the
effect is absent (and in domain D, inverted under the strict tier). The
exploratory findings (per-tier margin tests, pool gap, tension gradient) are labelled as such and
were not part of the frozen criteria; they motivate, but do not confirm,
the pre-registered follow-ups.

\paragraph{Pre-registered follow-ups}\label{sec:followups} (1) \emph{Diversity
dose--response:} hold the verifier fixed at S$^0$, manipulate pool
composition (temperature ladder, mixed LLM+random pools, multi-model),
frozen criterion: monotone holdout improvement in domain A with a domain-D
replication contrast. (2) \emph{Tension transfer:} manipulate capacity
utilization at fixed tier, frozen criterion: sign change of the S$^+$
group contrast near medium utilization, pre-registered before the next
unblinding. Criterion reachability is checked before freezing
(Section~\ref{sec:gate} lesson): a criterion that is structurally unsatisfiable under a
plausible degenerate outcome -- here S$^0\equiv$S$^-$ -- cannot
discriminate hypothesis from design.

\section{Limitations}
One proposer model (Qwen2.5-0.5B-Instruct); two domains; simulated
instances derived from real trajectories; the tension gradient varies
task density jointly with utilization (composition confound in the
declining right branch); the mix-pool analysis is exploratory (post-hoc)
and the union-pool promotion is derived from already-evaluated arm
promotions, not a pre-registered intervention; the
v1.0 strict-order criterion was structurally unsatisfiable under
S$^0\equiv$S$^-$, reported as a pre-registration design lesson. The
random-proposal arm is the only baseline: for the question \emph{does
verifier strength buy quality at a fixed proposer} this is the
appropriate control (any other proposer would confound pool composition
with verification strength), but it means the paper makes no claim about
absolute rule quality relative to human-designed or classical
heuristics; the domain-D mutation battery separates the tiers only
weakly (Section~5.1), so domain-A evidence carries the manipulation
claim; and all findings are established in two domains, with the
cross-domain generalization of the pool-composition lever left to the
pre-registered follow-ups. Three further scope limits: proposals within
an LLM session are strongly correlated, so the effective sample size of
the 100-pair comparisons is smaller than the pair count suggests; the
random arm spans only the frozen uniform weight family, so the
coverage-gap reading does not extend to other proposal distributions;
and the study is deliberately CPU-bound on commodity hardware, which
limits the scale -- though not the structure -- of what was evaluated.

\section{Conclusion}
Under frozen criteria across two real-data domains, verification
strength filters bad candidates on the mutation battery but does not buy
discovery quality on LLM proposals -- and in domain D it measurably
\emph{shrinks} the LLM-over-random advantage. The post-hoc evidence
identifies the composition of the proposer
pool as the more promising lever, now under pre-registered interventional
test, and the value of strictness is modulated by deployment tension.
A pre-registration design lesson is recorded: criteria must be checked
for reachability under plausible degenerate outcomes (here
S$^0\equiv$S$^-$) before freezing.
All ledgers, frozen criteria, mutation batteries, and reproduction
scripts are archived on Zenodo (DOI: 10.5281/zenodo.23308866).

\ack{We thank the maintainers of the public T-Drive dataset used in
domain~D.}

\dataccess{The pre-registered protocol (v1.0--v1.2 with changelogs), the
per-pair JSONL ledgers, the mutation batteries, the frozen verifiers, and
the seed-to-figure analysis scripts (including
\texttt{analyze\_tier1.py}, \texttt{final\_numbers.py},
\texttt{descriptive\_stats.py}, \texttt{mix\_pool\_analysis.py},
\texttt{floor\_sensitivity.py}, and \texttt{tension\_gradient.py}) are
archived on Zenodo under a timestamped DOI
(\url{https://doi.org/10.5281/zenodo.23308866}); the archived package is
available from the authors on reasonable request. Every quantitative result
is generated by these archived scripts; none is hand-computed. The software
environment is Python~3.10 with Hugging Face \texttt{transformers} on the
PyTorch CPU backend, run fully offline. The Beijing T-Drive dataset
\cite{yuan2010tdrive} is a public research dataset released by Microsoft
Research
(\url{https://www.microsoft.com/en-us/research/publication/t-drive-trajectory-data-sample/}),
used under its research-use terms; instance-generation scripts and frozen
instance manifests are part of the archived package, so instances can be
regenerated from the raw dataset rather than redistributed.}

\ethics{This study uses only the public, de-identified T-Drive trajectory
dataset; no personally identifying information, trip-purpose, or
vehicle-operator data are retained in the derived instances, and no new
human-subject data were collected. Automated research pipelines that act on
real infrastructure domains can propose rules whose deployment has safety
and energy consequences; our one-shot holdout semantics and frozen verifiers
are a partial mitigation, and we deploy no discovered rule. Negative results
in automated-research evaluation can be misread as endorsements of lax
verification; we emphasize the opposite reading -- verification is necessary
but its marginal value is regime-dependent, and pool composition deserves at
least equal design attention.}

\aucontribute{P.~C. and N.~Z. contributed equally to this work. P.~C. and
S.~Z. designed the study; S.~Z. implemented the pipeline, ran the experiments
and analyses, and drafted the manuscript; N.~Z. contributed to the
analysis and revision; all authors reviewed and approved the final version
and agree to be held accountable for the work performed therein.}

\funding{This research received no specific grant from any funding agency in
the public, commercial or not-for-profit sectors.}

\competing{The authors declare no competing interests.}

\disclaimer{During the preparation of this work the authors used a
large-language-model assistant to improve language and readability; the
authors reviewed and edited the content and take full responsibility for the
published article.}

\bibliographystyle{RS}
\bibliography{refs}

\begin{thebibliography}{99}

\bibitem{lu2024aiscientist}
Lu C, Lu C, Lange RT, Foerster J, Clune J, Ha D. 2024  The {AI} Scientist:
  Towards Fully Automated Open-Ended Scientific Discovery. {\em arXiv preprint
  arXiv:2408.06292}.

\bibitem{yuan2010tdrive}
Yuan J, Zheng Y, Zhang C, Xie W, Xie X, Sun G, Huang Y. 2010  {T-Drive}:
  Driving Directions Based on Taxi Trajectories. In {\em Proceedings of the
  18th SIGSPATIAL International Conference on Advances in Geographic
  Information Systems} pp. 99--108.

\bibitem{qwen25}
{Qwen Team}. 2024  Qwen2.5 Technical Report. {\em arXiv preprint
  arXiv:2412.15115}.

\bibitem{boiko2023coscientist}
Boiko DA, MacKnight R, Kline B, Gomes G. 2023  Autonomous Chemical Research
  with Large Language Models. {\em Nature} \textbf{624}, 570--578.

\bibitem{romera2024funsearch}
Romera-Paredes B, Barekatain M, Novikov A, Balog M, Kumar MP, Dupont E, Ruiz
  FJR, Ellenberg JS, Wang P, Fawzi O, Kohli P, Fawzi A. 2024  Mathematical
  Discoveries from Program Search with Large Language Models. {\em Nature}
  \textbf{625}, 468--475.

\bibitem{si2025ideas}
Si C, Yang D, Hashimoto T. 2025  Can {LLMs} Generate Novel Research Ideas? {A}
  Large-Scale Human Study with 100+ {NLP} Researchers. In {\em International
  Conference on Learning Representations}.

\bibitem{ding2026survey}
Ding T, Nannapaneni A, Liu B, Zhang L. 2026  Autonomous Research Agents: A
  Survey of {AI} Scientists and the Verification Gap. {\em arXiv preprint
  arXiv:2608.05179}.

\bibitem{zheng2023judging}
Zheng L, Chiang WL, Sheng Y, Zhuang S, Wu Z, Zhuang Y, Lin Z, Li Z, Li D, Xing
  EP, Zhang H, Gonzalez JE, Stoica I. 2023  Judging {LLM}-as-a-Judge with
  {MT-Bench} and Chatbot Arena. In {\em Advances in Neural Information
  Processing Systems} vol.~36 pp. 46595--46623.

\bibitem{cobbe2021verifiers}
Cobbe K, Kosaraju V, Bavarian M, Chen M, Jun H, Kaiser L, Plappert M, Tworek J,
  Hilton J, Nakano R, Hesse C, Schulman J. 2021  Training Verifiers to Solve
  Math Word Problems. {\em arXiv preprint arXiv:2110.14168}.

\bibitem{lightman2024lets}
Lightman H, Kosaraju V, Burda Y, Edwards H, Baker B, Lee T, Leike J, Schulman
  J, Sutskever I, Cobbe K. 2024  Let's Verify Step by Step. In {\em
  International Conference on Learning Representations}.

\bibitem{wang2023selfconsistency}
Wang X, Wei J, Schuurmans D, Le Q, Chi E, Narang S, Chowdhery A, Zhou D. 2023
  Self-Consistency Improves Chain of Thought Reasoning in Language Models. In
  {\em International Conference on Learning Representations}.

\bibitem{shinn2023reflexion}
Shinn N, Cassano F, Berman E, Gopinath A, Narasimhan K, Yao S. 2023  Reflexion:
  Language Agents with Verbal Reinforcement Learning. In {\em Advances in
  Neural Information Processing Systems} vol.~36 pp. 8634--8652.

\bibitem{madaan2023selfrefine}
Madaan A, Tandon N, Gupta P, Hallinan S, Gao L, Wiegreffe S, Alon U, Dziri N,
  Prabhumoye S, Yang Y, Gupta S, Majumder BP, Hermann K, Welleck S,
  Yazdanbakhsh A, Clark P. 2023  Self-Refine: Iterative Refinement with
  Self-Feedback. In {\em Advances in Neural Information Processing Systems}
  vol.~36 pp. 46534--46594.

\bibitem{stroebl2026limits}
Stroebl B, Kapoor S, Narayanan A. 2026  The Limits of Inference Scaling Through
  Resampling. In {\em International Conference on Learning Representations}.
arXiv:2411.17501.

\bibitem{brown2024monkeys}
Brown B, Juravsky J, Ehrlich R, Clark R, Le QV, R{\'e} C, Mirhoseini A. 2024
  Large Language Monkeys: Scaling Inference Compute with Repeated Sampling.
  {\em arXiv preprint arXiv:2407.21787}.

\bibitem{zhang2026measuring}
Zhang K, Li Y. 2026  Are We Measuring Scientific Intelligence? {Rethinking} the
  Evaluation of {AI} Scientists. {\em arXiv preprint arXiv:2610.04915}.
NeurIPS 2026 Workshop (Oral).

\bibitem{suwandi2026kernaut}
Suwandi RC, Yin F, Murphy K. 2026  Kernel Autoresearch for Open-Ended Model
  Discovery. {\em arXiv preprint arXiv:2610.10394}.

\bibitem{mouret2015elites}
Mouret JB, Clune J. 2015  Illuminating Search Spaces by Mapping Elites. .

\bibitem{pugh2016qd}
Pugh JK, Soros LB, Stanley KO. 2016  Quality Diversity: A New Frontier for
  Evolutionary Computation. {\em Frontiers in Robotics and {AI}} \textbf{3},
  40.

\bibitem{lehman2011novelty}
Lehman J, Stanley KO. 2011  Abandoning Objectives: Evolution Through the Search
  for Novelty Alone. {\em Evolutionary Computation} \textbf{19}, 189--223.

\bibitem{basu2026budget}
Basu A, Chakraborty P. 2026  Budget-Sensitive Discovery Scoring: A Formally
  Verified Framework for Evaluating {AI}-Guided Scientific Selection. {\em
  arXiv preprint arXiv:2603.12349}.

\bibitem{osc2015science}
{Open Science Collaboration}. 2015  Estimating the Reproducibility of
  Psychological Science. {\em Science} \textbf{349}, aac4716.

\bibitem{errington2021replicability}
Errington TM, Mathur M, Soderberg CK, Denis A, Perfito N, Iorns E, Nosek BA.
  2021  Investigating the Replicability of Preclinical Cancer Biology. {\em
  eLife} \textbf{10}, e71601.

\bibitem{gundersen2018state}
Gundersen OE, Kjensmo S. 2018  State of the Art: Reproducibility in Artificial
  Intelligence. In {\em Proceedings of the {AAAI} Conference on Artificial
  Intelligence} vol.~32 pp. 2317--2324.

\bibitem{pineau2021reproducibility}
Pineau J, Vincent-Lamarre P, Sinha K, Larivi{\`e}re V, Beygelzimer A,
  d'Alch{\'e} Buc F, Fox E, Larochelle H. 2021  Improving Reproducibility in
  Machine Learning Research: A Report from the {NeurIPS} 2019 Reproducibility
  Program. {\em Journal of Machine Learning Research} \textbf{22}, 1--20.

\bibitem{kapoor2023leakage}
Kapoor S, Narayanan A. 2023  Leakage and the Reproducibility Crisis in
  Machine-Learning-Based Science. {\em Patterns} \textbf{4}, 100804.

\end{thebibliography}

\end{document}